\documentclass[12pt]{iopart}

\usepackage{iopams} 
\usepackage{graphicx}
\usepackage{mdwmath}
\usepackage{tabularx}
\usepackage{multirow}
\usepackage{url}
\usepackage{setspace}
\usepackage{float}
\usepackage{cite}
\usepackage{setspace}
\usepackage{xcolor}
\usepackage{adjustbox}
\usepackage{color}
\usepackage{iopams}
\usepackage{algorithm2e}
\usepackage[T1]{fontenc}

\begin{document}
\maketitle
\title{SleepNet: Automated sleep analysis via dense convolutional neural network using physiological time series}

\author{B Pourbabaee$^1$, M H Patterson$^2$, M R Patterson$^3$ and F Benard$^4$}

\ead{\mailto{$^1$bpourbabaee@gmail.com, $^2$matthp@me.com, $^3$patterson.m@gmail.com and $^4$frederic.benard@gmail.com}}
\vspace{10pt}
\begin{indented}
\item[]February 2019
\end{indented}

\begin{abstract}
In this work, a dense recurrent convolutional neural network (DRCNN) was constructed to detect sleep disorders including arousal, apnea and hypopnea using Polysomnography (PSG) measurement channels provided in the 2018 Physionet challenge database. Our model structure is composed of multiple dense convolutional units (DCU) followed by a bidirectional long-short term memory (LSTM) layer followed by a softmax output layer. The sleep events including sleep stages, arousal regions and multiple types of apnea and hypopnea are manually annotated by experts which enables us to train our proposed network using a multi-task learning mechanism. Three binary cross-entropy loss functions corresponding to sleep/wake, target arousal and apnea-hypopnea/normal detection tasks are summed up to generate our overall network loss function that is optimized using the Adam method. Our model performance was evaluated using two metrics: the area under the precision-recall curve (AUPRC) and the area under the receiver operating characteristic curve (AUROC). To measure our model generalization, 4-fold cross-validation was also performed. For training, our model was applied to full night recording data. Finally, the average AUPRC and AUROC values associated with the arousal detection task were 0.505 and 0.922, respectively on our testing dataset. An ensemble of four models trained on different data folds improved the
AUPRC and AUROC to 0.543 and 0.931, respectively. Our proposed algorithm achieved first place in the official stage of the 2018 Physionet challenge for detecting sleep arousals with AUPRC of 0.54 on the blind testing dataset. 
\end{abstract}

\vspace{2pc}
\noindent{\it Keywords}: Sleep Arousal, Apnea-Hypopnea, Polysomnography, Dense Convolutional Neural Network, LSTM 

\section{Introduction: }
\label{Intro}
Sleep quality is critical for good health. Decreased sleep quality is associated with negative health outcomes such as depression \cite{tsuno2005sleep}, obesity \cite{cappuccio2008meta} and a higher risk of mortality due to cardiovascular diseases \cite{suzuki2009sleep}. Apnea and hypopnea are common sleep disorders which cause poor sleep quality \cite{engleman2004sleep}. A significant amount of research has gone into automated apnea/hypopnea events detection via machine learning techniques using Polysomnography (PSG) data \cite{pombo2017classification, otero2008fuzzy}, which is the standard method to investigate the sleep quality, and to detect any respiratory or non-respiratory related sleep disorders through measuring multiple physiological signals when the subject is asleep \cite{chesson1997indications}. Apnea/hypopnea events are relatively simple to detect within PSG data, this is reflected in the high inter-rater reliability that has been observed for their scoring \cite{magalang2013agreement}. Apnea and hypopnea events have been well researched and have been linked to multiple negative health outcomes. 

Sleep arousal due to factors other than apnea and hypopnea is another form of sleep disruption. Arousal can become an issue if it happens too often during sleep. According to the American Academy of Sleep Medicine (AASM) guidelines, arousal is an abrupt shift within Electroencephalogram (EEG) signal frequency bands, including alpha, theta and greater than 16 Hz which lasts at least 3 seconds and is preceded with at least 10 seconds of stable condition. During the rapid eye movement (REM) stage, the arousal may also appear with an increase in chin Electromyogram (EMG) signal \cite{halasz2004nature, berry2012aasm}. 

Sleep arousals due to factors other than apnea and hypopnea are a less researched form of sleep disruption. Non-apnea/hypopnea arousals can be respiratory effort related (RERA), or else, they may be due to teeth grinding, pain, bruxisms, hypo-ventilation, insomnia, muscle jerks, vocalizations, snores, periodic leg movement, Cheyne-Stokes breathing or respiratory obstructions that are not severe enough to be classified as apnea or hypopnea \cite{bonnet1986performance}. Normally, RERA is the most common type of non-apnea/hypopnea arousal. Very little research has been done concerning the effect that non-apnea/hypopnea arousals have on sleep quality and general health because they are costly and difficult to detect using traditional methods; manual detection of sleep arousals have been shown to have lower inter-scorer reliability when compared to apnea/hypopnea \cite{magalang2013agreement}. 

An automated method of detecting non-apnea/hypopnea arousals would be more cost-effective and less time consuming than traditional manual assessment by Polysomnography technicians. This would increase the pace of research in the area, as more data could be processed in less time at a lower cost. A well functioning, automated method of non-apnea/hypopnea arousal detection would be of benefit to health researchers in determining the effects that these events have on health as well as developing more effective treatments to reduce arousal frequency. 

Original work in this space utilized frequency analysis of EEG channels to automatically detect sleep arousal \cite{drinnan1996automated}. Other research compared decision tree methods, logistic regression methods and naive Bayes methods in the prediction of arousals, with the decision tree method having the highest accuracy and recall and the naive Bayes method having the highest precision \cite{espiritu2015automated}. Recent research used an adaptive threshold method on time and frequency features to automatically detect arousals from PSG data; this was shown to be significantly more reliable than between human raters \cite{coppieters2016automatic}.  Deep learning methods can model large data-sets better than heuristic algorithms or algorithms based on other machine learning methods such as decision trees or logistic regression that assume an underlying structure to the model. Different deep learning methods are used in the 2018 Physionet challenge to detect non apean/hypopnea sleep arousals \cite{Runnan2018Identification, varga2018using, thrainsson2018automatic, li2018sleep, shoeb2018evaluating, zabihi20191d}. A good review of all the methods and models that were designed for the challenge as well as the clinical and demographic characteristics of the challenge dataset are also given in \cite{ghassemi2018you}. 

The purpose of this work is to determine how accurately non-apnea/hypopnea arousals can be detected with the use of deep learning methods. Automating the accurate detection of sleep arousals may allow larger scale studies to be performed to better determine what future health risks are correlated with arousal frequency. There are a few research directions to consider when trying to convert our algorithm into something that is widely applicable. The first concern is that all of the data from this challenge was collected with one equipment setup and from one site. Extending this so that it can work with different sensor channel availabilities and ensuring that it preserves its performance across a range of different sites is still necessary. Moreover, although the advent of deep learning approaches has enable us to process big amount of biomedical data with less or no feature engineering process and to also improve the performance of our detection algorithms, the explainability and interpretability of the generated results to sleep physicians and professional clinicians is still a big challenge. The robustness, reproducibility and consistency of our designed deep learning approaches are other important requirements for having a positive impact on the future of sleep medicine. 
More discussions regarding the benefits and limitations of this study is also given in Section \ref{Discussion}. This work was done as part of the Physionet/Computing in Cardiology Challenge 2018 and is the extended version of \cite{Pourbabaee2018Automated}. 

\section{Methodology}
\label{methodology}
Recently, convolutional neural networks (CNN) have gained a lot of interest in physiological signal processing due to their ability to learn complex features in an end to end fashion without extracting any hand-crafted features \cite{pourbabaee2017deep, pourbabaee2018deep}. In this work, a dense recurrent convolutional neural network is proposed primarily to detect arousal regions as well as apnea/hypopnea and sleep/wake intervals using PSG data provided in the 2018 Physionet challenge. Our network is a modified DenseNet that is proposed in \cite{huang2017densely} and is composed of multiple dense convolutional units (DCU), where each is a sequence of convolutional layers that are all connected to provide maximum information flow. It ends with a bidirectional long-short term memory layer (LSTM) with a residual skip connection and extra convolutions to convert the LSTM hidden states from forward and backward passes to the output shape. To compute the probability of different sleep events at each sample during training process as well as computing losses, a remapping mechanism is also proposed to simplify the network decision making process. Moreover, other task labels such as apnea-hypopnea/normal and sleep/wake are used as auxiliary tasks in a multi-task learning framework to share representations between related tasks and to improve our model generalization on our desired task which is the arousal detection.

\section{Materials and Pre-Processing}
\label{Material}
The dataset includes PSG data from 1,985 subjects which were monitored at the MGH sleep laboratory for the diagnosis of sleep disorders. The data were partitioned in two sets; the first set (n = 994), and the second set (n = 989), where only the first set's labels were provided publicly to train and evaluate a model detecting target arousal intervals. In this work, the first set was divided into 794 training, 100 validation and 100 testing records, and the second set is the blind test set that was only used in the challenge to rank submitted models. Each record in this dataset includes multiple physiological signals that were all sampled at 200 Hz and were manually scored by certified sleep technicians at the MGH sleep laboratory according to the AASM guidelines. More details regarding the dataset and available annotations for different sleep analysis purposes are provided in \cite{ghassemi2018you}.

In this work, the PSG measurements (12 channels) are used to design an arousal detector model. The electrocardiogram (ECG) signal which is not necessary for sleep scoring is excluded from our analysis. First, an anti-aliasing finite impulse response (FIR) filter is applied to all channels, where the -3dB cut-off point is 28.29 Hz.  Second, the channels are downsampled to 50
Hz and the DC bias is removed. Finally, the channels are individually normalized by removing the mean and the root-mean-square (RMS) of every channel signal in a moving 18-minute window using fast Fourier transform (FFT) convolution which is the speed-optimized form of a regular convolution. According to the AASM guidelines, the baseline breathing is established in 2 minutes. Normalizing over 18-minute interval ensures 90\% overlap between the two ends of the baseline window. To make it clear, an example is given as follows: If the beginning of the 2-minute baseline window occurs at 9 minutes, then the end of the baseline window will occur at 11 minutes. The sliding window that will be used to normalize the sample at 9 minutes will be from 0 minutes to 18 minutes, and the sliding window that will be used to normalize the sample at 11 minutes will be from 2 minutes to 20 minutes. The total normalization window used is 20 minutes in duration, and 18 minutes is 90\% of this. A high percentage overlap is desired to make sure that any important variation in breathing is not normalized out. Our proposed normalization process is not applied to the oxygen saturation (SaO2) measurement that is only scaled to be limited in (-0.5, 0.5) to avoid saturating the neural network with large values.

\section{Sleep Disorder Detector Model}
\label{ModelStructure}
In this section, the DRCNN structure that is proposed to detect arousal regions as well as other sleep events is explained. The multi-task learning framework is also described in which all available annotations associated with the sleep/wake, arousal and apnea-hypopnea/normal events are employed to improve our network generalization. 

\subsection{DRCNN Network Structure}
\label{DrcnnStructure}
\begin{figure}
\centering
\includegraphics[width=0.8\columnwidth]{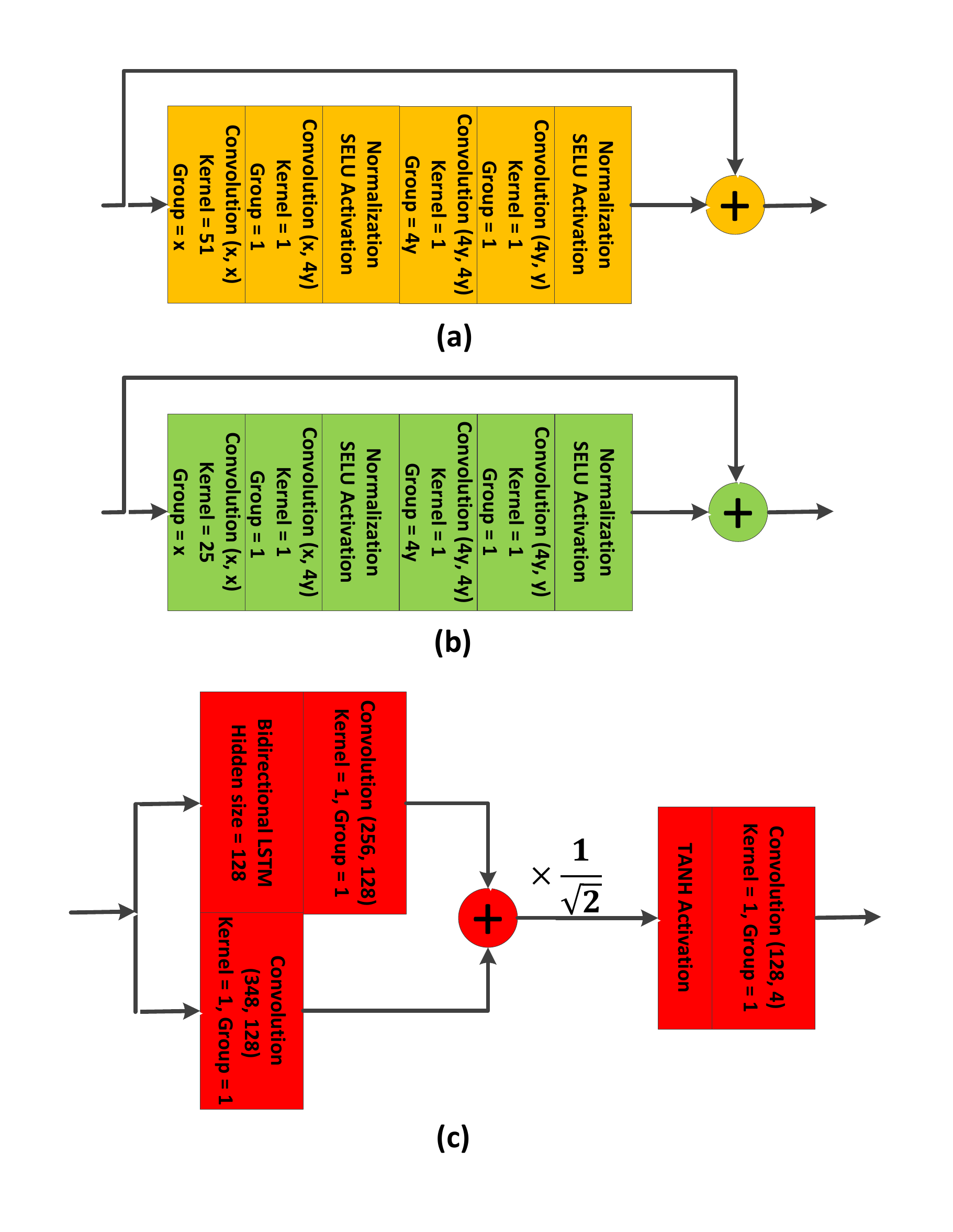}
\vspace{-12mm}
\caption{(a) DCU1, with no position-wise normalization, (b) DCU2, with position-wise normalization and (c) LSTM block, where $x$ and $y$ are the dimensions of input and output channels of DCUs. It must be noted that the "+" sign indicates the concatenation operation in DCU1 and DCU2 diagrams, but indicates the summation in the LSTM block.}
\label{blockDiagram2}
\end{figure}

\begin{figure*}
\centering
\includegraphics[width=1.0\columnwidth]{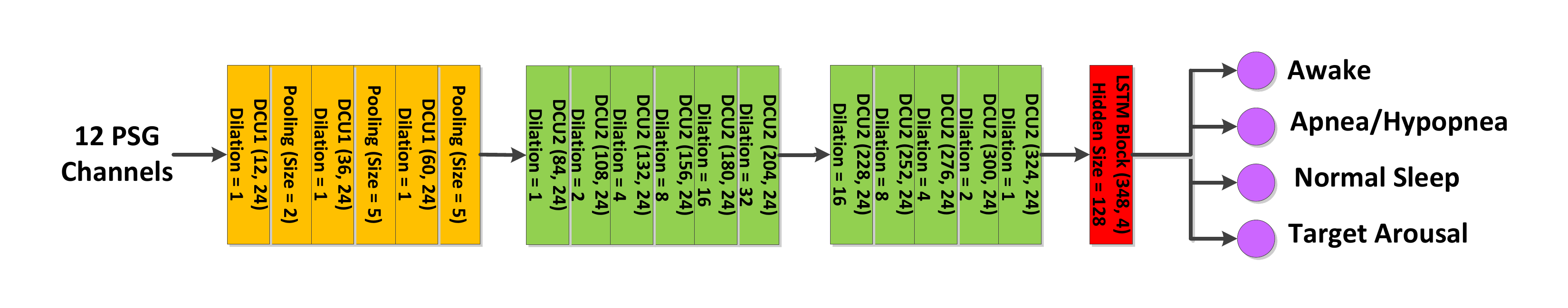}
\vspace{-8mm}
\caption{Proposed DRCNN architecture, including DCU1, DCU2 and LSTM block, where input and output channel dimensions are given in parentheses.}
\label{blockDiagram1}
\end{figure*}

In this work, our proposed DRCNN is trained and evaluated using data downsampled to 50 Hz to decrease computational effort and to fit a full night recording into memory to be applied to the network. The network is composed of multiple blocks, DCU1, DCU2 and LSTM which are displayed in \Fref{blockDiagram2}. First, there are three DCU1s, each followed by a max-pooling layer to down-sample input signals to one entity per second. According to \Fref{blockDiagram1}, the total pooling size of three successive max-pooling layers is $2*5*5 = 50$, which enables one to down-sample the 50 Hz input signal to 1 Hz. This is followed by eleven DCU2s. The DCU1s and DCU2s have similar structure comprising two sequences of two depthwise separable convolutional layers followed by the scaled exponential linear unit (SELU) activation functions. 

In DCU2, weight normalization, position-wise normalization and stochastic batch normalization \cite{dinh2016density} with a channel specific affine transform are also applied on convolutional layer outputs before using SELU activation function. Position-wise normalization involves subtracting the mean and dividing by the standard deviation across the channel dimension independently for each time step. To extend the DCU2 receptive field, dilated convolutions are also employed, where the dilation rates are first increased exponentially with the depth of the network along the first six DCU2s, and then are exponentially decreased along the remaining ones \cite{bai2018empirical}. However, in DCU1, neither a position-wise normalization nor a dilation factor is applied. Stochastic batch normalization is used in both DCU1 and DCU2.

Following the DCUs, an LSTM layer with a residual skip connection (linear $1\times1$ convolution) is also applied across the input channel temporal dimension. Finally, two more convolutional layers with $1\times1$ mapping are used to convert the LSTM hidden states from forward and backward passes to the output shape. The hyperbolic tangent (tanh) is applied before the last convolutional layer that leads to the more stable training process. Weight normalization is applied on each of the three convolutional layers in the LSTM block. The overall structure of our proposed DRCNN is displayed in \Fref{blockDiagram1}. The usefulness and contribution of each element of our proposed DRCNN structure is evaluated in details in \ref{ablation}, where the ablation study results are given.   

\subsection{Learning Mechanism}
\label{Learning Mechanism}
In this work, a multi-task learning mechanism is used to improve the generalization of our proposed arousal detector model and to learn more complex features through using other correlated tasks such as apnea-hypopnea/normal and sleep/wake detection. The ground truth corresponding to each task is a vector with two or three conditions that is defined as follows:
\begin{itemize}
\item Target arousal detection task: (target arousal = 1, non-target arousal (apnea/hypopnea or wake) = -1, and normal = 0),
\item Apnea-hypopnea/normal detection task: (all types of apnea/hypopnea = 1, and normal = 0),
\item Sleep/wake detection task: (sleep stages (REM, NREM1, NREM2, NREM3) = 1, wake = 0, and undefined stage = -1)
\end{itemize}

\begin{figure}
\centering
\includegraphics[width=\textwidth, trim={0 0 2in 0}]{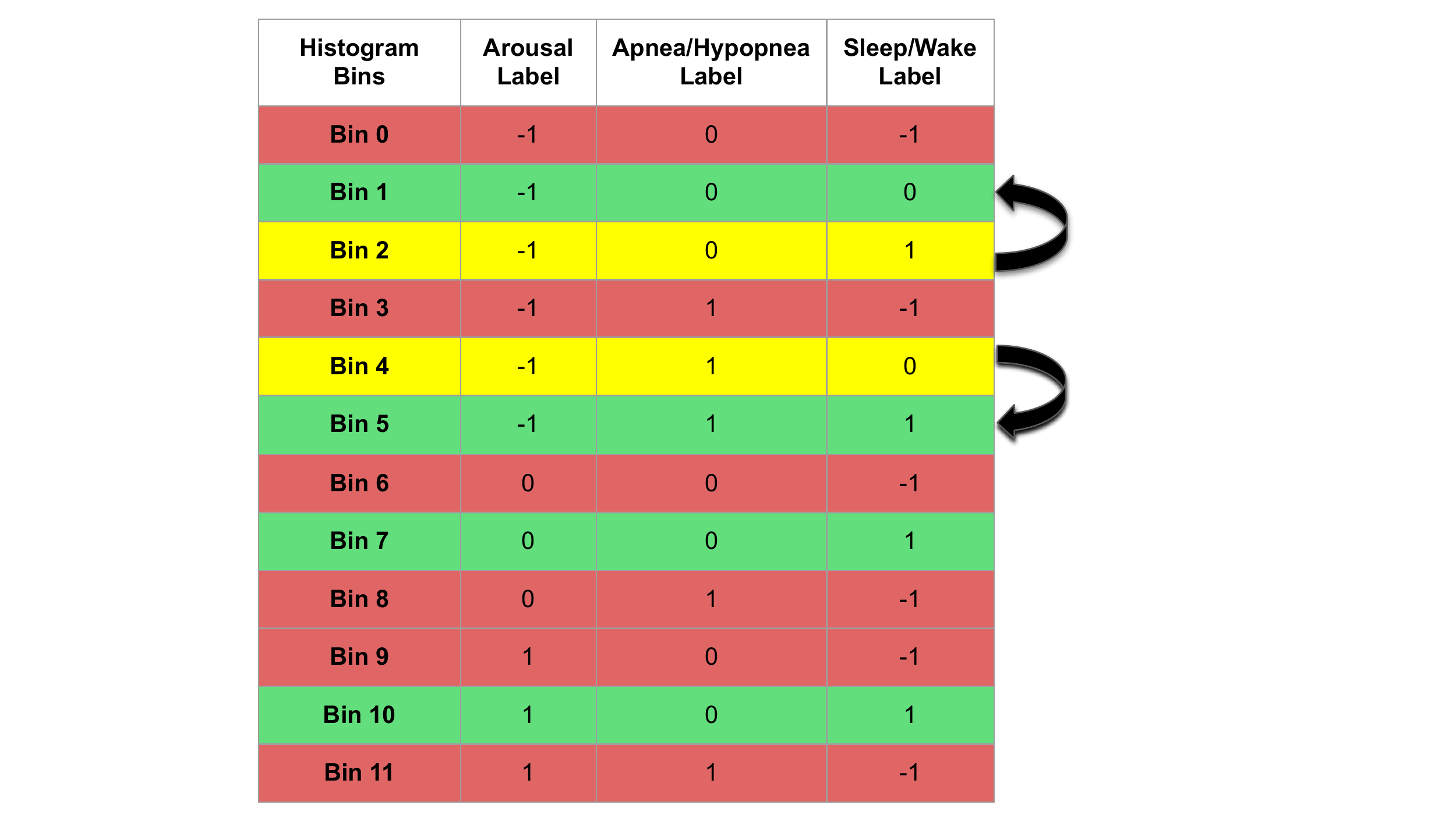}
\caption{Bins remapping mechanism to simplify multi-task learning process: The four green rows respectively indicate the sleep status (awake, apnea-hypopnea, normal sleep and target arousal) which constitutes the big portion of the data, the six red bins indicate the initial stage of the sleep which is undefined and constitute only a very small portion of the data for which the network is not penalized and the two yellow bins are the transition stages that are mapped to the green bins due to inconsistency among their labels. The definition of labels is given in the beginning of Section \ref{Learning Mechanism}.}
\label{remappingDiagram}
\end{figure}

Considering the above possible conditions associated with every task, 18 combinations can be defined. To investigate the distribution of the data associated with all combinations, a histogram of the labelled data was obtained. Among all 18 histogram bins, only 12 bins were non-empty. \Fref{remappingDiagram} displays 12 non-empty bins, where each represents a certain combination of labels corresponding to three detection tasks. To simplify the structure of the network output layer that computes joint probabilities, the non-empty bins are remapped to 4 bins that are displayed in green color in \Fref{remappingDiagram}. All the red bins corresponding to the beginning of the record before annotating the first sleep epoch (undefined sleep stage) are remapped to bin 0. The data associated with bin 0 are still processed by our model during training, however they do not contribute to the loss gradient. 

It is by definition impossible to get a sleep disorder while the subject is awake (condition in bin 4). This happens because according to the AASM guidelines, the sleep stages are annotated in 30-second epochs. Therefore, it is necessary to update sleep/wake detection task labels upon reaching such a state. For this purpose, bin 4 is remapped to bin 5. Similarly, bin 2 is remapped to bin 1 because when the arousal label is -1 and no apnea or hypopnea is present, the subject must be awake. 

The last convolutional layer of our proposed DRCNN has four output channels that are soft-maxed to compute joint probabilities corresponding to bin 1 (wakefulness), bin 5 (apnea-hypopnea), bin 7 (normal sleep) and bin 10 (target arousal). Then, the predicted target arousal, apnea-hypopnea/normal and sleep/wake marginal probabilities are computed as: P(target arousal) = P(bin 10), P(non-target arousal or normal) = P(bin 1) + P(bin 5) + P(bin 7), P(apnea/hypopnea) = P(bin 5), P(no apnea and hypopnea) = P(bin 1) + P(bin 7) + P(bin 10), P(wake) = P(bin 1), and P(sleep) = P(bin 5) + P(bin 7) + P(bin 10).

To train our DRCNN, the apnea-hypopnea/normal and sleep/wake are used as auxiliary detection tasks, whereas the target arousal detection is the desired task. The total cross-entropy loss is computed as the weighted average of loss values corresponding to the desired and auxiliary tasks, where the target arousal loss weight is set to 2 and the weights of other task losses are set to 1, since the auxiliary tasks are less important than the desired task. The higher target arousal loss weights did not have a positive impact on target arousal detection. The network weight parameters are optimized by using the Adam method without weight decaying. The learning rate is set to the default value of 0.001. In every epoch, 100 full-night recordings are randomly selected and processed through the network. The learning process is stopped if no improvement is achieved on validation records or no more decrease in validation loss. It must be noted that in our work, the epoch definition is different from the general notion; it is applying 100 full-night recordings to the network, not the full training set. Also, a record may appear in more than one epoch. Refer to the provided pseudo-code to get more information regarding our model learning process as well as the record indices corresponding to training, validation and testing data sets.

To evaluate the performance of the network, the AUPRC and AUROC are obtained for validation data and the model is checkpointed if there is any improvement with any of the above scores. The full training process is repeated four times across different folds of training and validation data and finally the predictions of our four models are averaged to obtain ensemble model predictions. The record indices corresponding to the four folds of cross validation are indicated in the following pseudo-code.

According to the given pseudo code, the shuffling of the indices uses a fixed random seed so that the splits are reproducible. Some details are omitted for clarity: SampleRecord() chooses a record index uniformly at random and loads the associated signals and targets, EvaluateValidationPerformance() computes the average precision for the arousal task from the model predictions over the validation data split. The testing split is not used in this procedure, but is shown for exposition. Note that we define an epoch as 100 updates of the model on 100 records, and one evaluation of the model on the validation data.

\begin{algorithm}
\SetAlgoLined
\KwResult{Trained Model}
 best\_validation\_average\_precision = 0\;
 num\_records\_per\_epoch = 100\;
 record\_indices = shuffle(0 to 994)\;
 fold\_number = 1\;
 \If{fold\_number == 1}{
   training\_indices = record\_indices[200:994]\;
    validation\_indices = record\_indices[100:200]\;
    testing\_indices = record\_indices[0:100]\;
   }
   \If{fold\_number == 2}{
   training\_indices = record\_indices[100:300;400:994]\;
    validation\_indices = record\_indices[300:400]\;
    testing\_indices = record\_indices[0:100]\;
   }
   \If{fold\_number == 3}{
   training\_indices = record\_indices[100:600;700:994]\;
    validation\_indices = record\_indices[600:700]\;
    testing\_indices = record\_indices[0:100]\;
   }
   \If{fold\_number == 4}{
   training\_indices = record\_indices[100:894]\;
    validation\_indices = record\_indices[894:994]\;
    testing\_indices = record\_indices[0:100]\;
   }
 \While{True}{
 \For{i in num\_records\_per\_epoch}{
 signals, targets = SampleRecord(training\_indices)\;
 predictions = model(signals)\;
 loss = ComputeLoss(predictions, targets)\;
 model = UpdateModel(model, loss)\;
 }
 validation\_average\_precision = EvaluateValidationPerformance(model, validation\_indices)\;
  \If{validation\_average\_precision \textgreater best\_validation\_average\_precision}{
   SaveModelCheckpoint(model)\;
   best\_validation\_average\_precision = validation\_average\_precision\;
   }
 }
\end{algorithm}

\section{Empirical Results}
\label{result}
The proposed DRCNN is applied to all available PSG channels, excluding ECG signal. The network hyper-parameters and learning procedure are explained in Section \ref{ModelStructure}. The PSG channels are first pre-processed as described in Section \ref{Material}. To train our network, the available annotated data are divided into four folds, where each includes 794 training, 100 validation and 100 consistent testing records. In each fold, a record only belongs to one of the training/validation/test dataset. Our network input is down-sampled to $50$ Hz and the input is composed of 12 PSG channels, each with the fixed duration of 7 hours (7*3600*50 samples). Shorter records are zero-padded, and the labels of the padded samples are set to (-1) which are all transferred to bin 0 in \Fref{remappingDiagram}. The data associated with bin 0 are still processed by our model during the training, however they do not contribute to the loss gradient. Using a multi-task learning process, the AUPRC and AUROC are obtained for sleep/wake, target arousal and apnea-hypopnea/normal detection tasks. \Tref{crossValidationResult} displays the performance metrics measured for each fold of cross-validation as well as the average performance on validation records across the 4 folds.
\begin{table}
\caption{Cross-validation results, where each model is evaluated on its own validation dataset.}
\vspace{2mm}
\centerline{\begin{adjustbox}{width=\columnwidth}
\begin{tabular}{lccccr} \hline\hline
Performance Metrics   & Model 1   & Model 2 & Model 3 & Model 4 & Average\\ \hline
Target Arousal AUROC & $0.922$ & $0.922$ & $0.913$ & $0.921$ & $0.919$ \\
Target Arousal AUPRC & $0.557$ & $0.505$ & $0.524$ & $0.529$ & $0.528$ \\
Apnea-Hypopnea/Normal AUROC  & $0.956$ & $0.958$ & $0.960$ & $0.972$ & $0.961$ \\
Apnea-Hypopnea/Normal AUPRC  & $0.734$ & $0.760$ & $0.764$ & $0.785$ & $0.760$ \\
Sleep/Wake AUROC      & $0.959$ & $0.958$ & $0.961$ & $0.937$ & $0.953$ \\
Sleep/Wake AUPRC     & $0.826$ & $0.834$ & $0.853$ & $0.767$ & $0.820$ \\ \hline\hline
\end{tabular}
\end{adjustbox}}
\label{crossValidationResult}
\end{table}

Using four trained models on different data folds, their corresponding predictions are averaged to form an ensemble model prediction. The ensemble model strategy improves the performance compared to the single model strategy. \Tref{ensembleModelResult} displays single and ensemble model performance evaluation results on the consistent test set. According to the details given in Section \ref{DrcnnStructure}, our network input and output frequencies are 50 Hz and 1 Hz , respectively. However, to evaluate the performance of our network on testing data set and to measure the performance metrics, the network output predictions are up-sampled to the original 200 Hz.
\begin{table}
\caption{Performance on testing records using single and ensemble model strategies.}
\vspace{2mm}
\centerline{\begin{adjustbox}{width=\columnwidth}
\begin{tabular}{lccccr} \hline\hline
Performance Metrics   & Model 1   & Model 2 & Model 3 & Model 4 & Ensemble\\ \hline
Target Arousal AUROC         & $0.921$ & $0.923$ & $0.923$ & $0.922$ & $0.931$ \\
Target Arousal AUPRC         & $0.492$ & $0.497$ & $0.519$ & $0.511$ & $0.543$ \\
Apnea-Hypopnea/Normal AUROC  & $0.951$ & $0.955$ & $0.954$ & $0.965$ & $0.965$ \\
Apnea-Hypopnea/Normal AUPRC  & $0.721$ & $0.745$ & $0.761$ & $0.781$ & $0.783$ \\
Sleep/Wake AUROC      & $0.958$ & $0.957$ & $0.958$ & $0.944$ & $0.960$ \\
Sleep/Wake AUPRC      & $0.831$ & $0.822$ & $0.822$ & $0.771$ & $0.832$ \\ \hline\hline
\end{tabular}
\end{adjustbox}}
\label{ensembleModelResult}
\end{table}

Finally, the average AUPRC and AUROC values associated with the target arousal detection task were $0.505$ and $0.922$, respectively on our testing dataset. An ensemble of four models trained on different data folds improved the AUPRC and AUROC to $0.543$ and $0.931$, respectively. To evaluate our ensemble network on other sleep events detection tasks, three popular metrics are measured as follows:
\begin{eqnarray}
     SE = \case{TST}{TRT} \\ 
     AI = \case{(Number \; of \; Arousals \; lasting \; more \; than \; 10 \; sec) \;  \times 60}{TST}\\ 
     AHI = \case{(Number \; of \; Apnea-Hypopnea \; lasting \; more \; than \; 10 \; sec) \; \times 60}{TST} 
\end{eqnarray}

\noindent where TST, TRT, SE, AI and AHI correspond to the total sleeping and recording times, sleep efficiency, arousal index and apnea-hypopnea index, respectively. According to the available sleep monitoring literature, the aforementioned metrics are used to identify subjects with sleep disorders as well as to estimate their severity. In \cite{nieto2000association}, the AHI is graded into four groups, namely as normal (AHI between 0 to 5), mild (AHI between 5 to 15), moderate (AHI between 15 to 30) and severe (AHI above 30). The higher AHI grades are the more serious sleep disorder problems which have to be treated appropriately using various methods such as the continuous positive airway pressure (CPAP) machine or other oral appliances. According to \cite{berry2012aasm}, it is required to measure the duration of an apnea-hypopnea condition to compute the AHI. Only the apnea-hypopnea episodes that are longer than 10 seconds contribute to the AHI computation. In our work, a label is predicted for every sample of an individual record. Hence, the duration of an apnea-hypopnea episode using the predicted labels corresponding to every sample of a record can easily be measured. To obtain the apnea-hypopnea predicted labels, a threshold of $0.2$ is applied on our model output prediction associated with the apnea-hypopnea. This threshold is set by trial and error, using only the training and validation data to maximize the apnea-hypopnea average detection accuracy. However, it would be recommended to incorporate this threshold in the training process to find its optimal value, which is part of the future work.

The mean absolute errors and the average actual and predicted values of the above metrics are measured and displayed in \Tref{SE_AI_AHI} for the first fold of the validation data as well as our testing records.
\begin{table}
\caption{Mean absolute error (MAE), average actual and predicted values of SE, AI and AHI measured for the first fold of validation set as well as testing records using the ensemble DRCNN model.}
\vspace{2mm}
\centerline{\begin{tabular}{lcc} \hline\hline
Performance Metrics & Validation & Test\\ \hline
SE MAE & $0.0534$ & $0.0612$ \\
AI MAE & $2.7988$ & $3.0705$ \\ 
AHI MAE & $4.1618$ & $5.0702$ \\
Actual Average SE & $0.8336$ & $0.8287$ \\
Predicted Average SE & $0.7844$ & $0.7741$ \\
Actual Average AI & $6.2972$ & $6.0031$ \\
Predicted Average AI & $6.2375$ & $6.4631$ \\
Actual Average AHI & $18.0809$ & $18.8664$ \\
Predicted Average AHI & $18.5142$ & $20.2659$ \\ \hline\hline
\end{tabular}}
\label{SE_AI_AHI}
\end{table}
The confusion matrix of the AHI grade estimation task using our DRCNN model is also displayed in Tables \ref{cm_AHI_validation} and \ref{cm_AHI_test} corresponding to validation and testing data sets, respectively. 
\begin{table}
 \centering
 \caption{Confusion matrix of apnea-hypopnea severity grade estimation task for the first fold of validation data using the ensemble DRCNN model.}
 \vspace{2mm}
 \begin{adjustbox}{width=\columnwidth}
 \begin{tabular}{lcccc} \hline\hline
 & Predicted Normal & Predicted Mild & Predicted Moderate & Predicted Severe \\ \hline
 Real Normal & $12$ & $2$ & $0$ & $0$ \\ 
 Real Mild & $4$ & $20$ & $3$ & $0$ \\ 
 Real Moderate & $0$ & $6$ & $35$ & $8$ \\ 
 Real Severe & $0$ & $0$ & $2$ & $8$ \\ \hline\hline
 \end{tabular}
 \end{adjustbox}
 \label{cm_AHI_validation}
\end{table}

\begin{table}
  \centering
 \caption{Confusion matrix of apnea-hypopnea severity grade estimation task for testing data set using the ensemble DRCNN model.}
 \vspace{2mm}
 \begin{adjustbox}{width=\columnwidth}
 \begin{tabular}{lcccc} \hline\hline
 & Predicted Normal & Predicted Mild & Predicted Moderate & Predicted Severe \\ \hline
 Real Normal & $12$ & $2$ & $0$ & $0$ \\ 
 Real Mild & $7$ & $16$ & $3$ & $0$ \\ 
 Real Moderate & $0$ & $12$ & $22$ & $14$ \\ 
 Real Severe & $0$ & $0$ & $0$ & $12$ \\ \hline\hline
 \end{tabular}
 \end{adjustbox}
 \label{cm_AHI_test}
\end{table}

To evaluate the performance of our model in estimating apnea-hypopnea severity grade, the accuracy, normal grade over-estimation rate (OSR) and the other grades under-estimation rates (USR) are computed and displayed in \Tref{PerformanceRate_AHIGrade} for validation and testing data sets using their corresponding confusion matrices. The normal grade OSR is the rate of subjects that are incorrectly diagnosed with higher apnea-hypopnea severity grades and the other grades USR is the rate of subjects within each category whose apnea-hypopnea severity grades are underestimated. 
\begin{table}
  \centering
 \caption{Overall accuracy, normal grade OSR as well as mild, moderate and severe apnea-hypopnea USRs for the first fold of validation data and the testing records using the ensemble DRCNN model.}
 \vspace{2mm}
 \begin{tabular}{lcc} \hline\hline
 Performance Metric & Validation & Test \\ \hline
 Overall Accuracy & $0.7500$ & $0.6200$ \\ 
 Normal Grade OSR & $0.1428$ & $0.1428$ \\
 Mild Grade USR & $0.1481$ & $0.2692$ \\ 
 Moderate Grade USR & $0.1224$ & $0.2500$ \\ 
 Severe Grade USR & $0.2000$ & $0.0000$ \\ \hline\hline
 \end{tabular}
 \label{PerformanceRate_AHIGrade}
\end{table}

The overall USR associated with all grades of apnea-hypopnea excluding the normal grade is $0.14$ and $0.22$ for the first fold of validation data and the testing records, respectively. Moreover, the sensitivity and specificity of our proposed method in estimating different apnea-hypopnea severity grades are measured and displayed in \Tref{AHIGrade_SenSpec}. Since the apnea-hypopnea severity grade is a multi-class problem, the sensitivity and specificity are calculated using a one-versus-all method.
\begin{table}
  \centering
 \caption{Sensitivity and specificity of our ensemble DRCNN model in estimating different apnea-hypopnea severity grades for the first fold of validation data and the testing records.}
 \vspace{2mm}
 \begin{tabular}{lcccc} \hline\hline
  & \multicolumn{2}{c}{Validation} & \multicolumn{2}{c}{Test} \\ 
  & Sensitivity & Specificity & Sensitivity & Specificity \\ \hline
   Normal & $0.857$ & $0.953$ & $0.857$ & $0.919$\\ 
   Mild & $0.741$ & $0.890$ & $0.615$ & $0.811$ \\
   Moderate & $0.714$ & $0.902$ & $0.458$ & $0.942$\\ 
   Severe & $0.800$ & $0.911$ & $1.000$ & $0.841$ \\ \hline\hline
 \end{tabular}
 \label{AHIGrade_SenSpec}
\end{table}

All experiments were performed on Nvidia GEFORCE GTX 1080 Ti GPU with the memory size of $11$ GB. The average training and validation times are $214$ and $110$ seconds, respectively. The code was developed in Python 3.6, using the Pytorch library and is shared on GitHub: \underline{https://github.com/matthp/Physionet2018\_Challenge\_Submission}.

\section{Discussion}
\label{Discussion}
According to the given empirical results in Section \ref{result}, the ensemble model strategy not only improves the target arousal detection performance metrics, but also enables us to detect apnea-hypopnea as well as sleep/wake intervals. Other researchers used deep learning as well as feature-based approaches in the 2018 Physionet challenge to detect non apean/hypopnea sleep arousals. Our work is compared with several of them in terms of the achieved AUPRC in \Tref{comparison}.
\begin{table}
  \centering
 \caption{Comparison of non-apnea/hypopnea sleep arousal detection on the Physionet 2018
Computing in Cardiology challenge.}
 \vspace{2mm}
 \begin{adjustbox}{width=\columnwidth}
 \begin{tabular}{lcc} \hline\hline
  Paper & Method & AUPRC \\ \hline 
  Shoeb \& Sridhar \cite{shoeb2018evaluating}$^*$ & Convolutional and recurrent neural network with model hyper-parmaters search & $0.573$ \\
  Our work & Dense convolutional neural network with LSTM & $0.543$ \\
  Varga et al. \cite{varga2018using} & Neural network with auxillary loss & $0.460$ \\
  Mar Priansson et al. \cite{thrainsson2018automatic} & Bidirectional recurrent neural network & $0.452$ \\
  He at al. \cite{Runnan2018Identification} & Deep neural networks with LSTM & $0.430$ \\
  Warrick \& Homsi \cite{warrick2018sleep} & Scattering transform and recurrent neural network & $0.375$ \\
  Li at al. \cite{li2018sleep} & End-to-end deep learning & $0.315$ \\
  Zabihi et al. \cite{zabihi20191d} & 1D convolutional neural network & $0.310$ \\
  Zabihi et al. \cite{zabihi2018automatic} & State distance analysis in phase space & $0.190$ \\ \\
  \multicolumn{3}{l}{(*) Submitted outside the time frame of the official stage of the 2018 Physionet challenge. The AUPRC is given for their internal test set, but not the official blind test set of the challenge.} \\ \hline\hline
 \end{tabular}
 \end{adjustbox}
 \label{comparison}
\end{table}

\Tref{comparison} shows that our proposed DRCNN outperforms the other models submitted to the 2018 Phsyionet challenge, excluding the one in \cite{shoeb2018evaluating} which utilized a hyper-parameter search to achieve the best model, but not during the official stage of the challenge. Having a hyper-parameter search is highly recommended as an extension of the current work to achieve the optimal results for the sleep arousal detection problem. Moreover, our model achieves a higher AUROC (0.931 vs. 0.916) on our test set compared to \cite{shoeb2018evaluating}. Additionally, the model in \cite{shoeb2018evaluating} is evaluated on an internal test set of 97 recordings (comparable to our testing dataset with a size of 100), but our model and the other given models in \Tref{comparison} are further validated on the 989 subjects in the blind test set.

Although our proposed DRCNN is primarily developed to detect arousal regions, the multi-task learning framework and the added auxiliary tasks enable us to deploy our model for detecting different types of sleep disorders including arousal, apnea and hypopnea. \Tref{SE_AI_AHI} confirms that our DRCNN model estimations of SE, AI and AHI are fairly accurate, thus can be used for generating the automated sleep monitoring report for sleeping subjects with low estimation errors. Note that in most of the misclassified cases, our model overestimated the apnea-hypopnea severity which is more acceptable than underestimating the severity grade or not detecting at all, since false positives are less onerous for a technician to correct than false negatives.

While not the main goal of this work, our model achieves reasonable performance on the detection of apnea/hypopnea events as well as sleep/wake conditions from PSG data. To have a fair comparison with other works on apnea/hypopne or sleep/wake detection, we need to consider either of them as the desired one (assigning a higher weight to its corresponding loss) or design a more focused network on the desired detection task. Therefore, it is possible that other existing available methods in the literature have achieved better results in apnea/hypopnea or sleep/wake detection task than what is achieved in this study.

In \cite{pittman2004assessment} a dynamic state modelling algorithm is utilized to automatically detect respiratory events (apena/hypopnea) from PSG data with an accuracy of 90\% compared to one annotator and 95\% to a second annotator. In \cite{xie2012real}, machine learning techniques are employed to detect sleep apnea in 25 patients with suspected sleep disordered breathing with accuracy, specificity and sensitivity all around 82\%. In \cite{yildirim2019deep}, a deep learning model is used to classify sleep/wake from PSG data with 98.06\% accuracy on the sleep-EDF public dataset \cite{goldberger2000physiobank} and 97.62\% accuracy on the expanded sleep-EDF data set. In \cite{khalighi2013automatic}, a support vector machine model is used to estimate sleep/wake stages from PSG data with an accuracy of 89.66\%, an F1 score of 83.25\%, specificity of 96.06\% and sensitivity of 83.26\%.

It is not clear what the clinical usefulness of this model is without further investigation. One area that needs to be clarified is a better comparison to human level performance. Each record in this dataset was only annotated by a single human scorer. Hence, comparison of the model outputs to the human annotations cannot account for inter-scorer variability \cite{magalang2013agreement}. A dataset with multiple independent annotations per record would provide an ability to compare the performance measures of different scorers against each other and against the model outputs. If this reveals a significant difference, than either more work needs to be done to bring this model to human level performance, or a better understanding of how an imperfect model can be usefully incorporated into the clinical workflow is necessary.

The primary limitation of this study is that the data is from a single site. There is no way to determine how a model trained on this data would generalize to other sites without collecting additional labeled data from those sites. This problem is further exacerbated by the dependence on 12 out of the 13 Polysomnography channels available from the equipment used at this site. Another site that does not use the same equipment, or at least provide the same channels, would not be able to use a trained model on this data. A new model with a similar architecture could be trained on data from different sites and equipment to develop a model that generalizes accordingly.

It must also be note that the low value for AUPRC ($\sim0.5$) and high value for AUROC ($\sim0.9$) indicates that there is more of a trade-off between sensitivity and precison than there is between sensitivity and specificity. For any given threshold, the sensitivity will be the same, but the specificity will be expected to be higher than the precision. This discrepancy between the two can occur in highly skewed datasets where most of the data is normal.

\section{Conclusion}
In this paper, a modified version of the dense convolutional neural network comprising multiple convolutional and LSTM blocks is proposed to detect sleep disorders including arousal, apnea and hypopnea using 12 PSG channels that are provided in the 2018 Physionet challenge database. To improve our network generalization and to use information from correlated tasks, a multi-task learning procedure using hard parameter sharing framework is also exploited in this work. Four DCRNN models are trained and evaluated on different subsets of training and validation data. Finally, an ensemble model is obtained through computing the average prediction of the above four models. The results confirm the superiority of the ensemble model against a single model approach. On the challenge blind testing dataset, the ensemble model achieves an AUPRC of $0.54$ which is the first-place entry in the Physionet challenge official stage.

\appendix
\section{Ablation Study}
\label{ablation}
To elucidate the contributions of our proposed DRCNN components, an extensive ablation study including multiple experiments was performed. In each experiment, only one component was modified or removed with respect to the baseline model that is our proposed DRCNN with the architecture given in \Fref{blockDiagram1}. All the ablation study models were trained and evaluated using the first fold of our data. The performance metrics of AUPRC and AUROC were measured for the first fold of validation data as well as the consistent testing records and were compared among different experiments. Tables \ref{ablationStudy_List} and \ref{ablationStudy_Validation} respectively give the list of ablation study experiments and the AUPRC and AUROC corresponding to all three tasks addressed in our multi-task learning framework for the first fold of the validation data set. Similarly, \Tref{ablationStudy_Testing} displays the ablation study results for our consistent testing records, using the model that is trained on the first fold. It must be noted that the training process was stopped at Epoch 500 in every ablation study, and the model with the highest performance on the validation data set was saved to be evaluated on our testing records.
\begin{table}
\caption{List of ablation study experiments.}
\vspace{2mm}
\centerline{\begin{adjustbox}{width=\columnwidth}
\begin{tabular}{lc} \hline\hline
Ablation Study & Experiment Description\\ \hline
Exp. 1 & Original Model (Our proposed DRCNN) \\
Exp. 2 & Replaced SELU activation functions with RELU \\ 
Exp. 3 & Turn on position-wise normalization in DCU1 block \\
Exp. 4 & Turn off position-wise normalization in DCU2 block \\
Exp. 5 & Remove bidirectional LSTM layer \\
Exp. 6 & Remove FFT convolution-based signal normalization explained in Section \ref{Material} \\ 
Exp. 7 & Remove residual mapping from the LSTM block \\
Exp. 8 & Remove weight normalization for all convolutional and LSTM layers \\
Exp. 9 & Remove auxiliary tasks (apply single task (arousal detection) learning) \\
Exp. 10 & Fix the dilation rates for the second part of the DCU2 block (last five dense units) to one \\ \hline\hline
\end{tabular}
\end{adjustbox}}
\label{ablationStudy_List}
\end{table}

\begin{table}
\caption{Ablation study results for the first fold of validation data set.}
\vspace{2mm}
\centerline{\begin{adjustbox}{width=\columnwidth}
\begin{tabular}{lcccccc} \hline\hline
Ablation Study &  Arousal AUROC & Arousal AUPRC & Apnea-Hypopnea/Normal AUROC & Apnea-Hypopnea/Normal AUPRC & Sleep/Wake AUROC & Sleep/Wake AUPRC\\ \hline
Exp. 1 & $0.923$ & $0.565$ & $0.955$ & $0.744$ & $0.958$ & $0.810$ \\
Exp. 2 & $0.930$ & $0.575$ & $0.959$ & $0.756$ & $0.958$ & $0.829$ \\
Exp. 3 & $0.924$ & $0.559$ & $0.956$ & $0.744$ & $0.959$ & $0.832$ \\
Exp. 4 & $0.918$ & $0.547$ & $0.951$ & $0.726$ & $0.945$ & $0.767$ \\
Exp. 5 & $0.927$ & $0.567$ & $0.956$ & $0.748$ & $0.962$ & $0.828$ \\
Exp. 6 & $0.924$ & $0.565$ & $0.956$ & $0.740$ & $0.960$ & $0.836$ \\
Exp. 7 & $0.540$ & $0.084$ & $0.582$ & $0.104$ & $0.558$ & $0.284$ \\
Exp. 8 & $0.920$ & $0.552$ & $0.953$ & $0.718$ & $0.955$ & $0.820$ \\
Exp. 9 & $0.919$ & $0.553$ & $0.586$ & $0.101$ & $0.640$ & $0.238$ \\
Exp. 10 & $0.925$ & $0.563$ & $0.956$ & $0.740$ & $0.959$ & $0.830$ \\ \hline\hline
\end{tabular}
\end{adjustbox}}
\label{ablationStudy_Validation}
\end{table}

\begin{table}
\caption{Ablation study results for the testing records using the single model strategy that is trained on the first data fold.}
\vspace{2mm}
\centerline{\begin{adjustbox}{width=\columnwidth}
\begin{tabular}{lcccccc} \hline\hline
Ablation Study &  Arousal AUROC & Arousal AUPRC & Apnea-Hypopnea/Normal AUROC & Apnea-Hypopnea/Normal AUPRC & Sleep/Wake AUROC & Sleep/Wake AUPRC\\ \hline
Exp. 1 & $0.923$ & $0.510$ & $0.952$ & $0.737$ & $0.959$ & $0.823$ \\
Exp. 2 & $0.927$ & $0.517$ & $0.955$ & $0.746$ & $0.958$ & $0.845$ \\
Exp. 3 & $0.919$ & $0.487$ & $0.953$ & $0.734$ & $0.960$ & $0.838$\\
Exp. 4 & $0.915$ & $0.474$ & $0.946$ & $0.711$ & $0.941$ & $0.757$ \\
Exp. 5 & $0.927$ & $0.522$ & $0.953$ & $0.741$ & $0.958$ & $0.827$ \\
Exp. 6 & $0.924$ & $0.511$ & $0.953$ & $0.739$ & $0.959$ & $0.826$ \\
Exp. 7 & $0.519$ & $0.071$ & $0.572$ & $0.107$ & $0.568$ & $0.294$ \\
Exp. 8 & $0.918$ & $0.497$ & $0.950$ & $0.718$ & $0.955$ & $0.830$ \\
Exp. 9 & $0.915$ & $0.491$ & $0.600$ & $0.107$ & $0.654$ & $0.253$ \\
Exp. 10 & $0.922$ & $0.508$ & $0.953$ & $0.736$ & $0.960$ & $0.836$ \\ \hline\hline
\end{tabular}
\end{adjustbox}}
\label{ablationStudy_Testing}
\end{table}

\begin{figure}
\centering
\vspace{-5mm}
\includegraphics[width=0.6\textwidth]{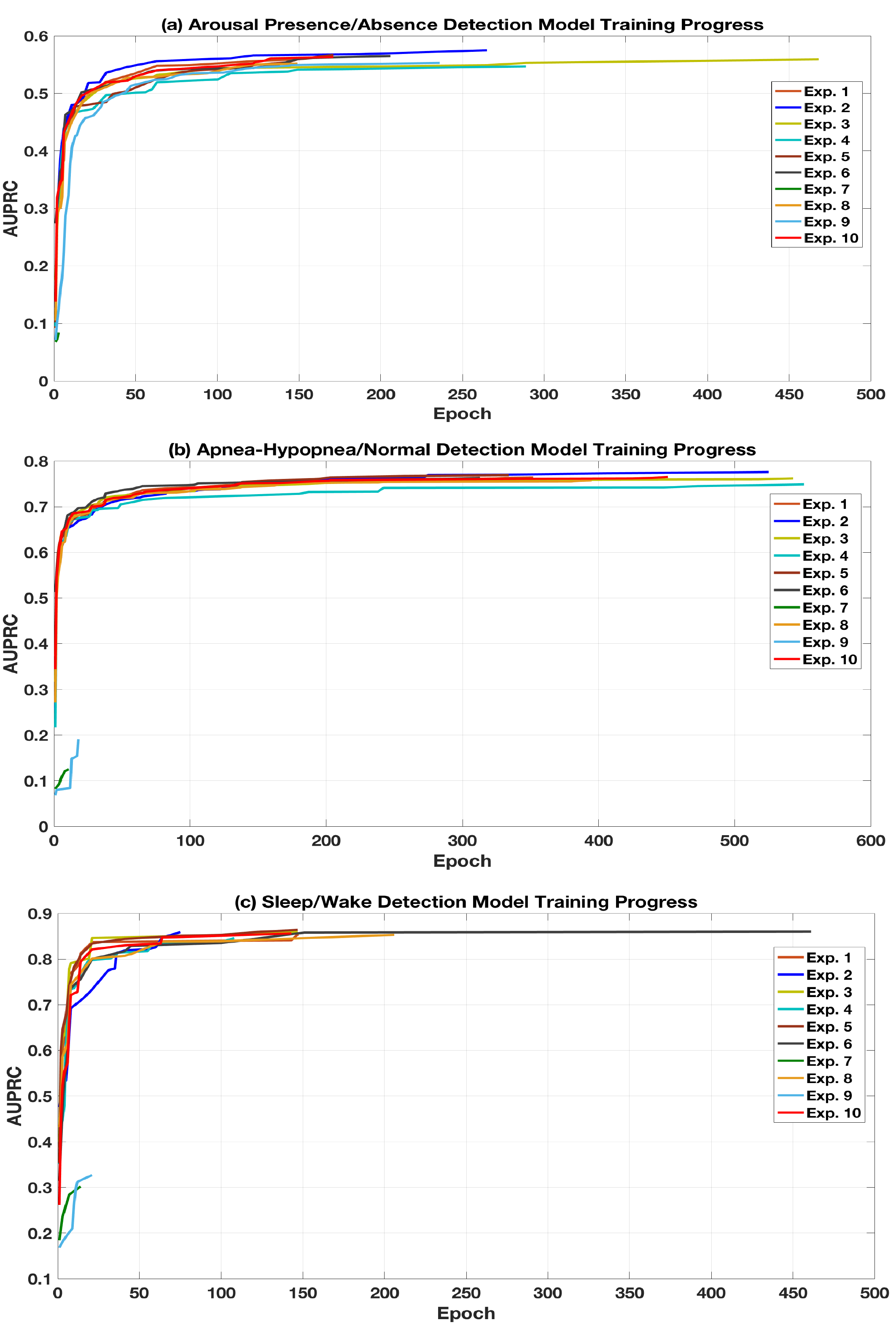}
\caption{Training progress of models, given in Table \ref{ablationStudy_List}, measured by the improving AUPRC values of the (a) target arousal, (b) apnea-hypopnea/normal, and (c) sleep/wake detection tasks for the first fold of the validation dataset versus the epoch number.}
\label{convergencePlots}
\end{figure}

It can be concluded from the ablation study models applied on our testing records, that the AUPRC and AUROC performance metrics are marginally or highly decreased in all experiments, excluding Exp. 2 and Exp. 5, compared to our proposed original model. This confirms the positive contribution of the components that were added to our proposed structure, specifically the contribution of the residual mapping in the LSTM block as well as activating the position-wise normalization in DCU2 and disabling it in DCU1. 

In experiments 2 and 5, the AUPRC and AUROC are slightly improved as compared to the original model, which is not a major issue but still need further investigations. It seems that both RELU and SELU activation functions work similarly in this problem, however SELU is still preferred over RELU due to its self-normalizing benefit that limits the risk of dying neurons \cite{klambauer2017self}. 

In order to compare the convergence speed of the experiments, the training progress is displayed in \Fref{convergencePlots} associated with our three detection tasks of every ablation study, where the improving AUPRC values are depicted versus the corresponding epoch number. According to \Fref{convergencePlots}, the original model converged to its highest AUPRC value on the validation set faster that the other models, excluding the model trained in Exp. 2. The only model that is as good as the original one in terms of the convergence speed is the model trained in Exp. 10 (with fixed dilation) which is not as accurate as the original model in terms of the AUPRC values. Although the AUPRC values obtained from the model in Exp. 5 are marginally higher than those obtained from the original model, the original model converges faster. It must also be noted that the AUPRC/AUROC values as well as the convergence trends of the apnea-hypopnea/normal and sleep/wake detection tasks obtained from the model in Exp. 9 (single task) are not valid, because the model was only trained to detect arousal. 

As a result, considering both the performance metrics and the training convergence speed, our proposed original model outperforms the other models that are evaluated in the ablation study, except the Exp. 2 (SELU replaced by RELU). To evaluate the contribution of the SELU activation function, further experiments are suggested to be performed using other training and validation data folds as well as the blind testing set which is out of the scope of the current paper.

\section*{References}
\bibliography{IEEEabrv,PhysMeasuJournal}

\begin{thebibliography}{10}
\providecommand{\url}[1]{#1}
\csname url@samestyle\endcsname
\providecommand{\newblock}{\relax}
\providecommand{\bibinfo}[2]{#2}
\providecommand{\BIBentrySTDinterwordspacing}{\spaceskip=0pt\relax}
\providecommand{\BIBentryALTinterwordstretchfactor}{4}
\providecommand{\BIBentryALTinterwordspacing}{\spaceskip=\fontdimen2\font plus
\BIBentryALTinterwordstretchfactor\fontdimen3\font minus
  \fontdimen4\font\relax}
\providecommand{\BIBforeignlanguage}[2]{{%
\expandafter\ifx\csname l@#1\endcsname\relax
\typeout{** WARNING: IEEEtran.bst: No hyphenation pattern has been}%
\typeout{** loaded for the language `#1'. Using the pattern for}%
\typeout{** the default language instead.}%
\else
\language=\csname l@#1\endcsname
\fi
#2}}
\providecommand{\BIBdecl}{\relax}
\BIBdecl

\bibitem{tsuno2005sleep}
N.~Tsuno, A.~Besset, and K.~Ritchie, ``Sleep and depression,'' \emph{The
  Journal of clinical psychiatry}, 2005.

\bibitem{cappuccio2008meta}
F.~P. Cappuccio, F.~M. Taggart, N.-B. Kandala, A.~Currie, E.~Peile,
  S.~Stranges, and M.~A. Miller, ``Meta-analysis of short sleep duration and
  obesity in children and adults,'' \emph{Sleep}, vol.~31, no.~5, pp. 619--626,
  2008.

\bibitem{suzuki2009sleep}
E.~Suzuki, T.~Yorifuji, K.~Ueshima, S.~Takao, M.~Sugiyama, T.~Ohta,
  K.~Ishikawa-Takata, and H.~Doi, ``Sleep duration, sleep quality and
  cardiovascular disease mortality among the elderly: a population-based cohort
  study,'' \emph{Preventive medicine}, vol.~49, no. 2-3, pp. 135--141, 2009.

\bibitem{engleman2004sleep}
H.~Engleman and N.~Douglas, ``Sleep{\textperiodcentered} 4: Sleepiness,
  cognitive function, and quality of life in obstructive sleep apnoea/hypopnoea
  syndrome,'' \emph{Thorax}, vol.~59, no.~7, pp. 618--622, 2004.

\bibitem{pombo2017classification}
G.~N. Pombo, Nuno and K.~Bousson, ``Classification techniques on computerized
  systems to predict and/or to detect apnea: A systematic review,''
  \emph{Computer methods and programs in biomedicine}, vol. 140, pp. 265--274,
  2017.

\bibitem{otero2008fuzzy}
A.~Otero, P.~Felix, M.~R. Alvarez, and C.~Zamarron, ``Fuzzy structural
  algorithms to identify and characterize apnea and hypopnea episodes,'' in
  \emph{30th Annual International Conference of the IEEE Engineering in
  Medicine and Biology Society (EMBS)}, 2008, pp. 5242--5245.

\bibitem{chesson1997indications}
J.~R. Chesson, L.~Andrew, R.~A. Ferber, J.~M. Fry, M.~Grigg-Damberger, K.~M.
  Hartse, T.~D. Hurwitz, S.~Johnson, G.~A. Kader, M.~Littner, G.~Rosen
  \emph{et~al.}, ``The indications for polysomnography and related
  procedures,'' \emph{Sleep}, vol.~20, no.~6, pp. 423--487, 1997.

\bibitem{magalang2013agreement}
U.~J. Magalang, N.-H. Chen, P.~A. Cistulli, A.~C. Fedson, T.~G{\'\i}slason,
  D.~Hillman, T.~Penzel, R.~Tamisier, S.~Tufik, G.~Phillips \emph{et~al.},
  ``Agreement in the scoring of respiratory events and sleep among
  international sleep centers,'' \emph{Sleep}, vol.~36, no.~4, pp. 591--596,
  2013.

\bibitem{halasz2004nature}
P.~Hal{\'a}sz, M.~Terzano, L.~Parrino, and R.~B{\'o}dizs, ``The nature of
  arousal in sleep,'' \emph{Journal of Sleep Research}, vol.~13, no.~1, pp.
  1--23, 2004.

\bibitem{berry2012aasm}
R.~B. Berry, C.~L. Albertario, S.~Harding, R.~M. Lioyd, D.~T. Plante, S.~F.
  Quan, M.~M. Troester, and B.~V. Vaughn, ``The {AASM} manual for the scoring
  of sleep and associated events,'' \emph{Rules, Terminology and Technical
  Specifications, Version 2.5, American Academy of Sleep Medicine}, p. 2018.

\bibitem{bonnet1986performance}
M.~H. Bonnet, ``Performance and sleepiness as a function of frequency and
  placement of sleep disruption,'' \emph{Psychophysiology}, vol.~23, no.~3, pp.
  263--271, 1986.

\bibitem{drinnan1996automated}
M.~Drinnan, A.~Murray, J.~White, A.~Smithson, C.~Griffiths, and G.~Gibson,
  ``Automated recognition of eeg changes accompanying arousal in respiratory
  sleep disorders,'' \emph{Sleep}, vol.~19, no.~4, pp. 296--303, 1996.

\bibitem{espiritu2015automated}
H.~Espiritu and V.~Metsis, ``Automated detection of sleep disorder-related
  events from polysomnographic data,'' in \emph{IEEE International Conference
  on Healthcare Informatics}, 2015, pp. 562--569.

\bibitem{coppieters2016automatic}
D.~Coppieters’t~Wallant, V.~Muto, G.~Gaggioni, M.~Jaspar, S.~L. Chellappa,
  C.~Meyer, G.~Vandewalle, P.~Maquet, and C.~Phillips, ``Automatic artifacts
  and arousals detection in whole-night sleep eeg recordings,'' \emph{Journal
  of neuroscience methods}, vol. 258, pp. 124--133, 2016.

\bibitem{Runnan2018Identification}
R.~He, K.~Wang, Y.~Liu, N.~Zhao, Y.~Yuan, Q.~Li, and H.~Zhang, ``Identification
  of arousals with deep neural networks using different physiological
  signals,'' \emph{Computing in Cardiology, Maastricht, Netherlands.}, vol.~45,
  2018.

\bibitem{varga2018using}
B.~Varga, M.~G{\"o}r{\"o}g, and P.~Hajas, ``Using auxiliary loss to improve
  sleep arousal detection with neural network,'' \emph{Computing in Cardiology,
  Maastricht, Netherlands.}, vol.~45, 2018.

\bibitem{thrainsson2018automatic}
H.~M. {\TH}r{\'a}insson, H.~Ragnarsd{\'o}ttir, G.~F. Kristj{\'a}nsson, and
  B.~Marin{\'o}sson, ``Automatic detection of target regions of respiratory
  effort-related arousals using recurrent neural networks,'' \emph{Computing in
  Cardiology, Maastricht, Netherlands.}, vol.~45, 2018.

\bibitem{li2018sleep}
H.~Li, Q.~Cao, Y.~Zhong, and Y.~Pan, ``Sleep arousal detection using end-to-end
  deep learning method based on multi-physiological signals,'' \emph{Computing
  in Cardiology, Maastricht, Netherlands.}, vol.~45, 2018.

\bibitem{shoeb2018evaluating}
A.~Shoeb and N.~Sridhar, ``Evaluating convolutional and recurrent neural
  network architectures for respiratory-effort related arousal detection during
  sleep,'' \emph{Computing in Cardiology, Maastricht, Netherlands.}, vol.~45,
  2018.

\bibitem{zabihi20191d}
M.~Zabihi, A.~B. Rad, S.~Kiranyaz, S.~S{\"a}rkk{\"a}, and M.~Gabbouj, ``1d
  convolutional neural network models for sleep arousal detection,''
  \emph{arXiv preprint arXiv:1903.01552}, 2019.

\bibitem{ghassemi2018you}
M.~M. Ghassemi, B.~E. Moody, L.~Lehman, C.~Song, Q.~Li, H.~Sun, R.~G. Mark,
  M.~B. Westover, and G.~D. Clifford, ``You snooze, you win: the
  physionet/computing in cardiology challenge 2018,'' \emph{Computing in
  Cardiology}, vol.~45, pp. 1--4, 2018.

\bibitem{Pourbabaee2018Automated}
M.~Howe-Patterson, B.~Pourbabaee, and F.~Benard, ``Automated detection of sleep
  arousals from polysomnography data using a dense convolutional neural
  network,'' \emph{Computing in Cardiology, Maastricht, Netherlands.}, vol.~45,
  2018.

\bibitem{pourbabaee2017deep}
B.~Pourbabaee, M.~Javan~Roshtkhari, and K.~Khorasani, ``Deep convolutional
  neural networks and learning {ECG} features for screening paroxysmal atrial
  fibrillation patients,'' \emph{IEEE Transactions on Systems, Man, and
  Cybernetics: Systems}, no.~99, pp. 1--10, 2017.

\bibitem{pourbabaee2018deep}
B.~Pourbabaee, M.~Howe-Patterson, E.~Reiher, and F.~Benard, ``Deep
  convolutional neural network for {ECG}-based human identification,'' vol.~41,
  2018.

\bibitem{huang2017densely}
G.~Huang, Z.~Liu, L.~Van Der~Maaten, and K.~Q. Weinberger, ``Densely connected
  convolutional networks,'' vol.~1, no.~2, 2017, pp. 4700--4708.

\bibitem{dinh2016density}
L.~Dinh, J.~Sohl-Dickstein, and S.~Bengio, ``Density estimation using real
  {NVP},'' \emph{arXiv preprint arXiv:1605.08803}, p. 2016.

\bibitem{bai2018empirical}
S.~Bai, J.~Z. Kolter, and V.~Koltun, ``An empirical evaluation of generic
  convolutional and recurrent networks for sequence modeling,'' \emph{arXiv
  preprint arXiv:1803.01271}, p. 2018.

\bibitem{nieto2000association}
F.~J. Nieto, T.~B. Young, B.~K. Lind, E.~Shahar, J.~M. Samet, S.~Redline, R.~B.
  D'agostino, A.~B. Newman, M.~D. Lebowitz, T.~G. Pickering \emph{et~al.},
  ``Association of sleep-disordered breathing, sleep apnea, and hypertension in
  a large community-based study,'' \emph{Jama}, vol. 283, no.~14, pp.
  1829--1836, 2000.

\bibitem{warrick2018sleep}
P.~Warrick and M.~N. Homsi, ``Sleep arousal detection from polysomnography
  using the scattering transform and recurrent neural networks,''
  \emph{Computing in Cardiology, Maastricht, Netherlands.}, vol.~45, 2018.

\bibitem{zabihi2018automatic}
M.~Zabihi, A.~B. Rad, S.~S{\"a}rkk{\"a}, S.~Kiranyaz, A.~K. Katsaggelos, and
  M.~Gabbouj, ``Automatic sleep arousal detection using state distance analysis
  in phase space,'' \emph{Computing in Cardiology, Maastricht, Netherlands.},
  vol.~45, 2018.

\bibitem{pittman2004assessment}
S.~D. Pittman, M.~M. MacDonald, R.~B. Fogel, A.~Malhotra, K.~Todros, B.~Levy,
  A.~B. Geva, and D.~P. White, ``Assessment of automated scoring of
  polysomnographic recordings in a population with suspected sleep-disordered
  breathing,'' \emph{Sleep}, vol.~27, no.~7, pp. 1394--1403, 2004.

\bibitem{xie2012real}
B.~Xie and H.~Minn, ``Real-time sleep apnea detection by classifier
  combination,'' \emph{IEEE Transactions on Information Technology in
  Biomedicine}, vol.~16, no.~3, pp. 469--477, 2012.

\bibitem{yildirim2019deep}
O.~Yildirim, U.~B. Baloglu, and U.~R. Acharya, ``A deep learning model for
  automated sleep stages classification using {PSG} signals,''
  \emph{International journal of environmental research and public health},
  vol.~16, no.~4, p. 599, 2019.

\bibitem{goldberger2000physiobank}
A.~L. Goldberger, L.~A. Amaral, L.~Glass, J.~M. Hausdorff, P.~C. Ivanov, R.~G.
  Mark, J.~E. Mietus, G.~B. Moody, C.-K. Peng, and H.~E. Stanley, ``Physiobank,
  physiotoolkit, and physionet: components of a new research resource for
  complex physiologic signals,'' \emph{Circulation}, vol. 101, no.~23, pp.
  e215--e220, 2000.

\bibitem{khalighi2013automatic}
S.~Khalighi, T.~Sousa, G.~Pires, and U.~Nunes, ``Automatic sleep staging: A
  computer assisted approach for optimal combination of features and
  polysomnographic channels,'' \emph{Expert Systems with Applications},
  vol.~40, no.~17, pp. 7046--7059, 2013.

\bibitem{klambauer2017self}
G.~Klambauer, T.~Unterthiner, A.~Mayr, and S.~Hochreiter, ``Self-normalizing
  neural networks,'' \emph{Advances in Neural Information Processing Systems},
  pp. 971--980, 2017.

\end{thebibliography}
\end{document}